\documentclass[conference]{IEEEtran}
\IEEEoverridecommandlockouts
 
\usepackage{cite}
\usepackage{amsmath,amssymb,amsfonts}
\usepackage{algorithmic}
\usepackage{graphicx}
\usepackage{subcaption}
\usepackage{standalone}
\usepackage{hyperref}
\usepackage{textcomp}
\def\BibTeX{{\rm B\kern-.05em{\sc i\kern-.025em b}\kern-.08em
    T\kern-.1667em\lower.7ex\hbox{E}\kern-.125emX}}

\usepackage{tikz}
\usepackage{pgfplots}
\pgfplotsset{compat=newest}
\usetikzlibrary{er, positioning, calc}

\usepackage{multirow}
\usepackage{xcolor,colortbl}
\usepackage{xcolor}

\definecolor{darkWhite}{rgb}{0.96,0.96,0.96}
\definecolor{bluekeywords}{rgb}{0.13,0.13,1}
\definecolor{greencomments}{rgb}{0,0.5,0}
\definecolor{redstrings}{rgb}{0.9,0,0}

\definecolor{Comment}{RGB}{97,161,176}

\definecolor{btfGreen}{RGB}{51,160,44}
\definecolor{btfRed}{RGB}{190,60,90}

\definecolor{bleuUni}{RGB}{0, 157, 224}
\definecolor{marronUni}{RGB}{68, 58, 49}

\definecolor{bluecite}{HTML}{009DE0}

\definecolor{Paired_1}{RGB}{31,120,180}
\definecolor{Paired_2}{RGB}{166,206,227}
\definecolor{Paired_3}{RGB}{51,160,44}
\definecolor{Paired_4}{RGB}{178,223,138}
\definecolor{Paired_5}{RGB}{227,26,28}
\definecolor{Paired_6}{RGB}{251,154,153}
\definecolor{Paired_7}{RGB}{255,127,0}
\definecolor{Paired_8}{RGB}{253,191,111}
\definecolor{Paired_9}{RGB}{106,61,154}
\definecolor{Paired_10}{RGB}{202,178,214}
\definecolor{Paired_11}{RGB}{177,89,40}
\definecolor{Paired_12}{RGB}{255,255,153}
\definecolor{Accent_1}{RGB}{127,201,127}
\definecolor{Accent_2}{RGB}{190,174,212}
\definecolor{Accent_3}{RGB}{253,192,134}
\definecolor{Accent_4}{RGB}{255,255,153}
\definecolor{Accent_5}{RGB}{56,108,176}
\definecolor{Accent_6}{RGB}{240,2,127}
\definecolor{Accent_7}{RGB}{191,91,23}
\definecolor{Accent_8}{RGB}{102,102,102}
\definecolor{Spectral_1}{RGB}{158,1,66}
\definecolor{Spectral_2}{RGB}{213,62,79}
\definecolor{Spectral_3}{RGB}{244,109,67}
\definecolor{Spectral_4}{RGB}{253,174,97}
\definecolor{Spectral_5}{RGB}{254,224,139}
\definecolor{Spectral_6}{RGB}{255,255,191}
\definecolor{Spectral_7}{RGB}{230,245,152}
\definecolor{Spectral_8}{RGB}{171,221,164}
\definecolor{Spectral_9}{RGB}{102,194,165}
\definecolor{Spectral_10}{RGB}{50,136,189}
\definecolor{Spectral_11}{RGB}{94,79,162}
\definecolor{Set1_1}{RGB}{228,26,28}
\definecolor{Set1_2}{RGB}{55,126,184}
\definecolor{Set1_3}{RGB}{77,175,74}
\definecolor{Set1_4}{RGB}{152,78,163}
\definecolor{Set1_5}{RGB}{255,127,0}
\definecolor{Set1_6}{RGB}{255,255,51}
\definecolor{Set1_7}{RGB}{166,86,40}
\definecolor{Set1_8}{RGB}{247,129,191}
\definecolor{Set1_9}{RGB}{153,153,153}
\definecolor{Set1_10}{RGB}{0,0,0}
\definecolor{Set2_1}{RGB}{102,194,165}
\definecolor{Set2_2}{RGB}{252,141,98}
\definecolor{Set2_3}{RGB}{141,160,203}
\definecolor{Set2_4}{RGB}{231,138,195}
\definecolor{Set2_5}{RGB}{166,216,84}
\definecolor{Set2_6}{RGB}{255,217,47}
\definecolor{Set2_7}{RGB}{229,196,148}
\definecolor{Set2_8}{RGB}{179,179,179}
\definecolor{Dark2_1}{RGB}{27,158,119}
\definecolor{Dark2_2}{RGB}{217,95,2}
\definecolor{Dark2_3}{RGB}{117,112,179}
\definecolor{Dark2_4}{RGB}{231,41,138}
\definecolor{Dark2_5}{RGB}{102,166,30}
\definecolor{Dark2_6}{RGB}{230,171,2}
\definecolor{Dark2_7}{RGB}{166,118,29}
\definecolor{Dark2_8}{RGB}{102,102,102}
\definecolor{Reds_1}{RGB}{255,245,240}
\definecolor{Reds_2}{RGB}{254,224,210}
\definecolor{Reds_3}{RGB}{252,187,161}
\definecolor{Reds_4}{RGB}{252,146,114}
\definecolor{Reds_5}{RGB}{251,106,74}
\definecolor{Reds_6}{RGB}{239,59,44}
\definecolor{Reds_7}{RGB}{203,24,29}
\definecolor{Reds_8}{RGB}{165,15,21}
\definecolor{Reds_9}{RGB}{103,0,13}
\definecolor{Greens_1}{RGB}{247,252,245}
\definecolor{Greens_2}{RGB}{229,245,224}
\definecolor{Greens_3}{RGB}{199,233,192}
\definecolor{Greens_4}{RGB}{161,217,155}
\definecolor{Greens_5}{RGB}{116,196,118}
\definecolor{Greens_6}{RGB}{65,171,93}
\definecolor{Greens_7}{RGB}{35,139,69}
\definecolor{Greens_8}{RGB}{0,109,44}
\definecolor{Greens_9}{RGB}{0,68,27}
\definecolor{Blues_1}{RGB}{247,251,255}
\definecolor{Blues_2}{RGB}{222,235,247}
\definecolor{Blues_3}{RGB}{198,219,239}
\definecolor{Blues_4}{RGB}{158,202,225}
\definecolor{Blues_5}{RGB}{107,174,214}
\definecolor{Blues_6}{RGB}{66,146,198}
\definecolor{Blues_7}{RGB}{33,113,181}
\definecolor{Blues_8}{RGB}{8,81,156}
\definecolor{Blues_9}{RGB}{8,48,107}

\begin{document}

\title{Federated learning compression designed for lightweight communications\\\thanks{This work is supported by the \textit{Futur et Ruptures} program funded by IMT and Institut Carnot TSN, and by the GdR ISIS.}}

\author{
    \IEEEauthorblockN{
    Lucas Grativol\IEEEauthorrefmark{1,2},
    Mathieu Léonardon\IEEEauthorrefmark{1},
    Guillaume Muller\IEEEauthorrefmark{3}, 
    Virginie Fresse\IEEEauthorrefmark{2} and
    Matthieu Arzel\IEEEauthorrefmark{1}}
    
    \IEEEauthorblockA{\IEEEauthorrefmark{1}IMT Atlantique, Lab-STICC, UMR CNRS 6285, F-29238 Brest, France}
    \IEEEauthorblockA{\IEEEauthorrefmark{2}Hubert Curien Laboratory, Saint-Etienne, France}
    \IEEEauthorblockA{\IEEEauthorrefmark{3}Mines Saint-Etienne, Institut Henri Fayol, Saint-Etienne, France}
}

\maketitle

\begin{abstract}
Federated Learning (FL) is a promising distributed method for edge-level machine learning, particularly for privacy-sensitive applications such as those in military and medical domains, where client data cannot be shared or transferred to a cloud computing server. In many use-cases, communication cost is a major challenge in FL due to its natural intensive network usage. Client devices, such as smartphones or Internet of Things (IoT) nodes, have limited resources in terms of energy, computation, and memory. To address these hardware constraints, lightweight models and compression techniques such as pruning  and quantization are commonly adopted in centralised paradigms. In this paper, we investigate the impact of compression techniques on FL for a typical image classification task. Going further, we demonstrate that a straightforward method can compresses messages up to 50\% while having less than 1\% of accuracy loss, competing with state-of-the-art techniques.
\end{abstract}


\section{Introduction}

\begin{figure}
    \centering
    \includegraphics[scale=0.78]{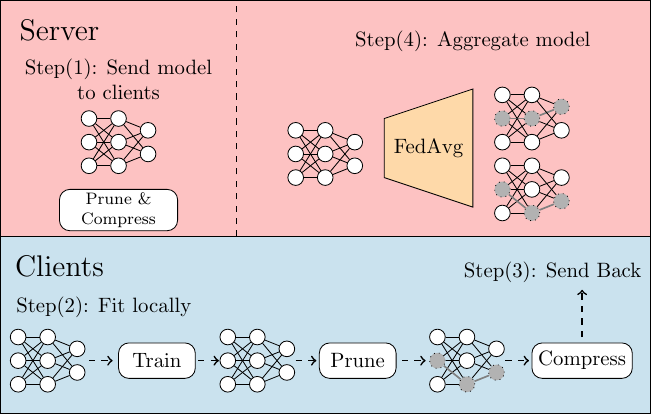}
    \caption{The pipeline of our study. We propose a simple way to insert the pruning technique as extra step before communicating training results.}
    \label{fig:prune_framework}
\end{figure}

The development of approaches for training machine learning models while preserving data privacy has long been a goal. In traditional machine learning, embedded systems send their raw data over a network to a powerful server, which then trains the model and sends it back. However, this process raises confidentiality issues, such as data interception during communication and unauthorised access to user data by the server owner or a third party. In standard Federated Learning (FL), the server sends a model to a group of clients, who train it on their local data and then send their updated parameters back to the server for aggregation. By reversing the training process in this way, FL attempts to better guarantee the confidentiality of user data, since data never leaves a client device. An overview of the process can be seen in Fig.~\ref{fig:prune_framework}.

These embedded devices such as IoT devices, smartphones and drones are well suited to FL applications due to their proximity to real-world data and applications~\cite{ray2021review}. However, many of these devices have limited computational resources and co-design techniques~\cite{cheng2017survey} are continually being explored to match algorithms to hardware constraints. Among the emerging research topics for FL at the edge/device level, the field of neural network compression is a promising way to tackle the constraints of devices exploiting FL~\cite{kairouz2021advances}.

In addition to message compression, the FL domain encompasses important ongoing research efforts. These include addressing challenges related to client heterogeneity in terms of both data and hardware~\cite{reddi2020adaptive}, ensuring secure aggregation against attacks~\cite{manzoor2022fedclamp}, and increasing client's privacy~\cite{kairouz2021advances}. While our work primarily focuses on reducing message sizes for energy and bandwidth reductions, we emphasize the importance of seamless integration with other ongoing research in FL. When compared to previous approaches~\cite{qiu2022zerofl,li2023anycostfl}, we propose a simpler and more effective solution that not only reduces message sizes but also ensures the possibility to be combined with other techniques without compromising accuracy. Our code is publicly accessible~\footnote{\url{https://github.com/lgrativol/fl_exps}}.

\section{Background}
\label{sec:background}

\subsection{Overview of Federated Learning}
\label{subsec:federated_learning}

Federated Learning (FL)~\cite{mcmahan2017communication} is a distributed framework that enables collaborative training of machine learning (ML) models on multiple devices, called clients, via a central coordinator, usually a server with large compute resources. Clients are commonly embedded devices, such as smartphones. Differing from traditional ML, each client trains its own local model and shares only the local training results, like model parameters or gradients, with the server. Through this mechanism, multiple clients  can jointly contribute to train a global model without sharing their data. In each federated training cycle, commonly referred to as a 'round,' the server distributes the current model to a subset of clients, who perform local training on the model and subsequently send back the updated results. The final step involves aggregating, on the server-side, client's contribution to create a global model, which ideally can represent the knowledge from each client. Each round involves downloading the model and several training iterations at client level.

\subsection{Model Compression}
\label{subsec:compression}

Model compression is a widely adopted solution~\cite{cheng2017survey} to reduce the computational and memory requirements of a model. Among existing compression techniques, quantization and pruning have been implemented to reduce the complexity of inference and training of neural networks~\cite{hoefler2021sparsity,lin2022device}.

Pruning aims to reduce the complexity of a model by removing redundant or unnecessary parts of an architecture. Very wide and deep models tend to yield good results, but the contribution of each of its elements to the performance of the whole network is not homogeneous. So, by observing each architectural element of the network, it is possible to eliminate those that have little impact. There are two possible approaches to pruning in the context of neural networks~\cite{tessier2022rethinking}. The first is to replace the value of certain weights with zeros, which is commonly referred to as unstructured pruning. The second approach consists of pruning entire structures within the network, such as kernels, filters or layers, which do not contribute significantly to the network's performance. This approach is known as structured pruning.  On the other hand, unstructured pruning can also offer compression benefits for FL through the use of entropy coding techniques, such as Huffman~\cite{han2015deep} coding, by exploiting sparse parameters. So far works in the literature~\cite{qiu2022zerofl} have used pruning to reduce communication cost by an order of 4.5 times. This is a significant consideration since typical FL clients often encompass low-power devices and operate in challenging transmission environments, such as long-distance or underwater communications.

Another widely applied technique is quantization, neural network models are generally constructed using 32-bit floating point numbers (FP32), which are more expensive in terms of computation, memory and energy than integers~\cite{horowitz20141}. In centralized machine learning is well-known that full-precision, FP32, it's not a necessary condition to obtain close to state-of-the-art results for inference and training~\cite{han2015deep}. 

ZeroFL~\cite{qiu2022zerofl} is a recent work that seeks to reduce simultaneously communication and training costs with a double optimization scheme to FL. First, a sparse training method named SWAT~\cite{raihan2020sparse}, and second, a layer-wise pruning based on weight importance. However as shown in~\cite{li2023anycostfl} communications cost can be much higher than the training cost. What should be done in the case where the focus is solely on communication costs ?

\section{Magnitude Pruning for Federated Learning}
\label{sec:meth}

We address the invoked problem in~\ref{sec:background} by proposing a distributed non-structured pruning method. Unlike previous works, our objective is to demonstrate that the conventional FL framework can be modified to support sparse messages. This method results in a compression of approximately 50\% of the original size while preserving accuracy with less than a 1\% loss. Our implementation is streamlined and easily extendable, making it compatible with more advanced FL algorithms.

Starting from the standard FL pipeline, we introduced pruning as a way to sparsify messages, server to client and vice-versa. Inspired by~\cite{han2015deep}, both server and clients perform a non-structured magnitude pruning just before transmitting a message. This pruning method is based on pruning the absolute value of the global weights following a predetermined pruning rate. Accordingly, the $\theta\%$ smaller weights are substituted with zeros, thereby pruned. Consequently, both the server and clients attain an equivalent level of sparsity in the message throughout each round.

At first, we conducted experiments to study the behaviour of our method while taking into account the impact on message compression. We applied different levels of pruning to detect trade-offs between compression and FL training mechanisms according to the experimental setup illustrated in Fig.~\ref{fig:prune_framework}. Building upon the results of our experiment, we extended our study to include a comparison with a recently published paper to showcase the viability of our approach.

\section{Experiments}
\label{sec:experiments}

\subsection{Exploring Magnitude Pruning}
\label{subsec:exp_prun}
To explore the impacts of our technique in FL, we simulated an image classification task, on the CIFAR-10 dataset, using a ResNet-12 with 780K parameters and 2.97 MB. We used the Flower~\cite{beutel2020flower} framework to simulate 10 FL clients. At each round, 40\% of the clients are selected for the training process, and 100 rounds were performed to study the evolution of the server model accuracy. Each client used SGD with momentum as the optimizer. For simplicity, we use the same hyperparameters as ~\cite{reddi2020adaptive}, also replacing the batch-norm layer by a group-norm layer. The server uses FedAvg as the aggregation strategy. 
Training examples are distributed across clients with a Latent Dirichlet Allocation (LDA)~\cite{hsu2019measuring} on the original training set. The LDA partition is controlled by a distribution parameter, $\alpha$. A smaller value of $\alpha$ results in a more non-IID task, making it more challenging. We examined the behavior of our technique in a relatively IID (Independent and Identically Distributed) scenario with $\alpha = 100$. In this setting, clients possess examples of all the classes.

\begin{figure}[!h]
    \centering
     \includegraphics[scale=0.72]{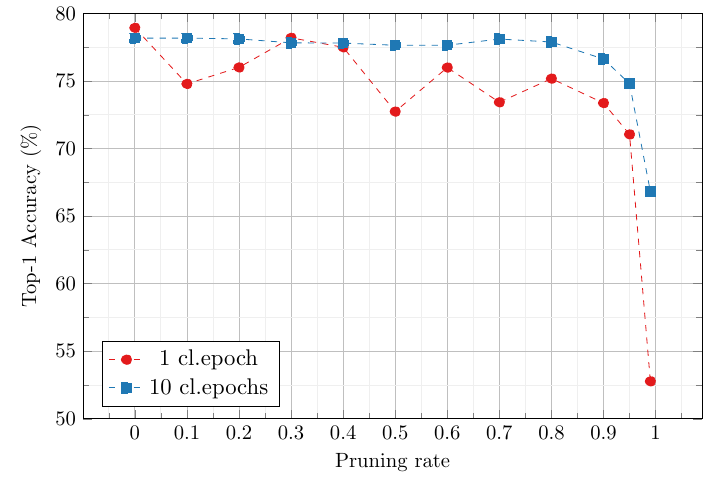}
     \caption{Pruning effect on the accuracy in function of the pruning rate, where the rate represents the \% of total parameters pruned, for 1 and 10 clients epochs.}
    \label{fig_prun}
\end{figure}

 As noted in previous works~\cite{mcmahan2017communication,kairouz2021advances} the number of local iterations performed by clients during training can have an important impact on model aggregation. For so, we decided to investigate this behaviour in the presence of model compression. The results on Fig.~\ref{fig_prun} show that spending more time on each client contributes to a more robust model, allowing sparser data communications while retaining approximately the same accuracy, even though this approach also results in a higher total number of local iterations.

To further investigate and evaluate the feasibility of our method in a non-IID scenario, we adopted the same test case as ZeroFL. The model is a ResNet-18 with 11M trainable parameters, occupying 44.7 MB. The FL scenario simulates 100 clients with 10\% participation rate, for only 1 local epoch and with $\alpha = 0.1$, where clients don't have access to all classes and the number of examples is randomly distributed. Table~\ref{tab_zerofl} presents the model evaluation results. The reported results are the means of three separate runs, with different seeds applied to generate distinct distributions of clients' data. Unless otherwise stated, the size of the models is reported after being compressed using a ZIP algorithm.

\begin{table}[]
\centering
\caption{Comparation to ZeroFL }
\label{tab_zerofl}
\begin{tabular}{|l|c|c|c|}
\hline
\multicolumn{1}{|c|}{Method} &  Compression &  \multicolumn{1}{c|}{Accuracy} &  
\begin{tabular}[c]{@{}c@{}}Message \\ Size (MB)\end{tabular} \\ \hline
\multicolumn{1}{|c|}{\multirow{3}{*}{\textbf{ZeroFL}}} &  \multicolumn{1}{c|}{\textbf{Full model}} &  \textbf{$80.62 \pm 0.72$} &  \textbf{44.7} \\ 
\multicolumn{1}{|c|}{} &  \cellcolor{Accent_3}\begin{tabular}[c]{@{}c@{}}90 \% SP + \\ 0.2 Mask Ratio\end{tabular} &  \multicolumn{1}{c|}{\cellcolor{Accent_3}$81.04\pm0.28$} &  \cellcolor{Accent_3}27.3 \\ 
\multicolumn{1}{|c|}{} &  \cellcolor{Spectral_8}\begin{tabular}[c]{@{}c@{}}90 \% SP + \\ 0.0 Mask Ratio\end{tabular} &  \cellcolor{Spectral_8}$73.87\pm0.50$ &  \cellcolor{Spectral_8}10.1 \\ 
\hline
\multirow{6}{*}{\textbf{\begin{tabular}[c]{@{}l@{}}Global \\ magnitude\\ (Ours)\end{tabular}}} & \multicolumn{1}{c|}{\textbf{Full model}} &  \textbf{$84.43\pm0.36$} & \textbf{44.7} \\
  & 10 \% pruning rate &  $85.96\pm0.37$ &  38.1 \\ 
  & 20 \% pruning rate &  $85.57\pm0.19$ &  34.8 \\ 
  & 30 \% pruning rate &  $85.03\pm0.32$ &  31.1 \\ 
  &  \cellcolor{Accent_3}40 \% pruning rate & \cellcolor{Accent_3}$85.20\pm0.20$ &  \cellcolor{Accent_3}27.1 \\ 
  & 50 \% pruning rate &  $83.85\pm0.65$ &  23.0 \\ 
  & 60 \% pruning rate &  $83.19\pm0.44$ &  18.9 \\ 
  & 70 \% pruning rate &  $82.25\pm0.63$ &  14.5 \\  
  &  \cellcolor{Spectral_8}80 \% pruning rate &   \cellcolor{Spectral_8}$80.70\pm0.24$ &   \cellcolor{Spectral_8}9.8 \\ 
  & 90 \% pruning rate &  $76.77\pm0.47$ &  4.9 \\ 
  & 95 \% pruning rate &  $69.14\pm0.85$ &  2.5 \\ 
  & 99 \% pruning rate &  $0.10 \pm 0.0$ &  0.5 \\ 
   \hline
\end{tabular}
\end{table}

In Table~\ref{tab_zerofl}, we present a comparison with ZeroFL~\cite{qiu2022zerofl}. Initially, without pruning, our baseline has a higher accuracy than ZeroFL and as far as we have understood, there are two main distinctions. Firstly, we do not employ SWAT for local training. Secondly, we use a batch size of 8, whereas ZeroFL does not indicate the specific batch size used. As previously observed in FL~\cite{mcmahan2017communication}, the batch size is a crucial hyperparameter that influences the aggregation accuracy. Even though SWAT plays a significant role in reducing the communication cost, it also has an impact on the model accuracy, resulting in an overall hindrance. This effect can be noticed as our baseline, which uses pure FedAvg without any compression, already achieves higher accuracy, 4\%, when compared to ZeroFL. We observe that for the same level of pruning, our approach exhibits proportionally less degradation. For instance, while ZeroFL experiences an 8\% accuracy degradation to prune the model to 10 MB, we only experience a 4.63\% degradation. As the results show, client's flexibility to perform pruning on its own better compensates for the sparsity introduced. This compensation enables messages to be more sparse while resulting in a more robust global model.

\subsection{Compressing more with Quantization}
\label{subsec:exp_quant}

\begin{figure}[!h]
\centering
    \includegraphics[scale=0.75]{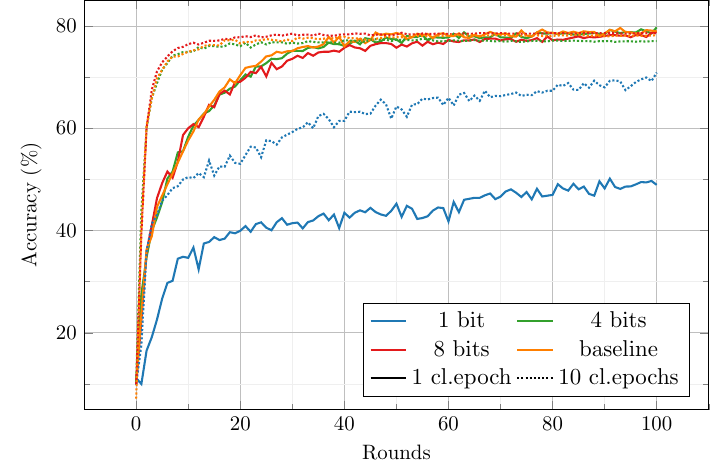}
    \caption{Accuracy evolution comparison between baseline (32-bit FP), 1-bit, 4-bit and 8-bit, for 1 and 10 clients epochs.}
    \label{fig_quant}
\end{figure}

From the message savings observed with the pruning experiments, one could ask if it is possible to have even smaller messages. As exposed in section~\ref{subsec:compression} another well-known technique for compression is quantization. In Fig~\ref{fig_quant}, we show the impact of Quantization-Aware Training (QAT)~\cite{alessandro_2021_4571063,lin2022device} in the IID scenario described before. During QAT, weights are still represented as floating-point numbers but are limited to power-of-two values. At each gradient update, the values are re-evaluated and scaled. The motivation behind using QAT is to incorporate quantization noise into the training procedure, allowing the network to learn from it. We chose to work with 1-bit, 4-bit and 8-bit quantization levels, using Binary Connect~\cite{courbariaux2015binaryconnect} for binary networks and the Brevitas~\cite{alessandro_2021_4571063} framework for 4- and 8-bit with the default quantization scheme. The weights are quantized to 4-bit and 8-bit integers and the QAT scaling is calculated per layer.

Looking at Fig.~\ref{fig_quant}, we can see that the convergence time, i.e. the number of rounds needed to reach maximum accuracy, is not the same from one experiment to another, as it also depends on the level of quantization. In addition to the fact that the 4- and 8-bit format enables us to achieve an accuracy comparable to the reference, it also reveals a compromise between communication and computation. In order to achieve a similar accuracy of around 75\%, Fig.~\ref{fig_quant}, it is necessary to perform 40 rounds of communication and 40 total epochs when using 1 local epoch, while in the case of 10 local epochs, 100 total epochs are needed within 10 rounds. Still, in the case of one bit, increasing the number of epochs per round on the client from 1 to 10 considerably increases accuracy, from 48.8~\% to 70.9~\%, with the total number of epochs increasing from 100 to 1000, with the same communication cost. As seen in the IID pruning experiment in Fig.~\ref{fig_prun}, spending more time on each client contributes to a more robust model to the perturbations introduced by the quantization.

\begin{table}[]
\centering
\caption{Summary of message size and accuracy for the CIFAR-10 dataset for the IID case}
\label{tab_iid}
\begin{tabular}{cccc}
\hline
\multicolumn{1}{c|}{\textbf{\begin{tabular}[c]{@{}c@{}}Compression\\ Technique\end{tabular}}} &
  \multicolumn{2}{c|}{\textbf{\begin{tabular}[c]{@{}c@{}}Accuracy \\ (\%)\end{tabular}}} &
  \textbf{\begin{tabular}[c]{@{}c@{}}Message Size \\ (MB)\end{tabular}} \\ \hline
\multicolumn{1}{l}{} &
  \begin{tabular}[c]{@{}c@{}}1 Local \\ Epoch\end{tabular} &
  \begin{tabular}[c]{@{}c@{}}10 Local \\ Epochs\end{tabular} &
  \multicolumn{1}{l}{} \\ \hline
Baseline  & 78.94 & 78.18 & 2.97 \\ \hline
Pruning &
  \multicolumn{1}{l}{} &
  \multicolumn{1}{l}{} &
  \multicolumn{1}{l}{} \\ \hline
10 \%     & 74.79 & 78.18 & 2.57 \\ 
20 \%     & 76.01 & 78.12 & 2.34 \\ 
30 \%     & 78.20 & 77.83 & 2.10 \\ 
40 \%     & 77.50 & 77.81 & 1.85 \\ 
50 \%     & 72.74 & 77.65 & 1.57 \\ 
60 \%     & 76.00 & 77.65 & 1.29 \\
70 \%     & 73.43 & 78.11 & 1.01 \\ 
80 \%     & 75.18 & 77.89 & 0.70 \\ 
90 \%     & 73.37 & 76.63 & 0.37 \\ 
95 \%     & 71.05 & 74.81 & 0.19 \\ 
99 \%     & 52.77 & 66.82 & 0.04 \\ \hline 
\multicolumn{4}{l}{Quantization} \\ \hline
8 bits    & 78.80 & 78.58 & 0.75 \\ 
4 bits    & 79.74 & 77.04 & 0.38 \\
1 bit     & 48.93 & 70.89 & 0.10 \\ \hline
\end{tabular}
\end{table}

Table~\ref{tab_iid} summarises the size of a message exchanged between client and server for IID scenario. For quantization, the message size depends only on the quantized weights, since the server knows the client's quantization. Table~\ref{tab_iid} also shows that even simple approaches can be used to compress a network, representing savings of 2 to 4 times in bandwidth without significantly affecting accuracy.

\section{Conclusion}

Federated learning represents a new approach to training models in a distributed manner, bringing forth fresh optimization challenges due to the presence of embedded systems serving as FL clients. These clients operate with limited hardware, energy, and communication resources. In this article, we demonstrated the promising application of traditional neural network compression methods in the context of FL. Our easy to implement yet effective technique achieved up to a 50\% reduction in message size without any significant impact on accuracy, thereby resulting in direct savings in energy and bandwidth costs. Moreover, our method allows each client to customize their pruning process, enabling greater flexibility to adapt to their unique datasets. By integrating quantization into the training process, we introduced an additional compression technique to the framework. It is conceivable that combining quantization and pruning could further enhance message compression, although our results already demonstrate the significance of both techniques individually. Based on these findings, we posit that incorporating a compression-aware training method, while ensuring seamless integration, is a crucial step in advancing the field of FL.

\bibliographystyle{IEEEtran}
\bibliography{IEEEabrv,main.bib}

\end{document}